# Revisiting Cross Modal Retrieval

Shah Nawaz[1], Muhammad Kamran Janjua[2], Alessandro Calefati[1], Ignazio Gallo[1]

University of Insubria[1], National University of Sciences and Technology[2],

**Abstract.** This paper proposes a cross modal retrieval system that leverages on image and text encoding. Most multimodal architectures employ separate networks for each modality to capture the semantic relationship between them. However, in our work image-text encoding can achieve comparable results in terms of cross modal retrieval without having to use separate network for each modality. We show that text encodings can capture semantic relationships between multiple modalities. In our knowledge, this work is the first of its kind in terms of employing a single network and fused image-text embedding for cross modal retrieval. We evaluate our approach on two famous multimodal datasets: MS-COCO and Flickr30K.

## 1 Introduction

Computer vision has started to pivot from classical image classification problems to associating text descriptions to visual data. The mapping of multiple modalities to a shared latent space has resulted in robust and exhaustive scene understanding. Due to these recent advancements [1–3], we see a surge in multimodal tasks such as image caption generation [4–8], visual question answering [9, 10], and audio-visual correspondence [11–14].

The problem we address in the paper is the multimodal image-text retrieval or simply bidirectional retrieval. The main target of the problem is to retrieve either modality at output given other modality at the input. For example in case of sentence-image retrieval, an image is retrieved given a sentence at input and vice versa [15, 16].

It is evident that neural network mappings are commonly used to bridge the gap between multiple modalities [17, 18] in building a joint representation of each modality. Typically, separate networks are trained to predict features of each modality and a supervision signal is employed to reduce the distance between image and associated text descriptions [19–22]. In addition, to capture text context before semantically associating it with the visual data, some techniques employ RNNs [23, 24] along with CNNs stacked in a CRNN fashion [25–27].

The major drawback of these techniques is that they rely on raw image-text pairs. However, if two different images have two similar text descriptions, the

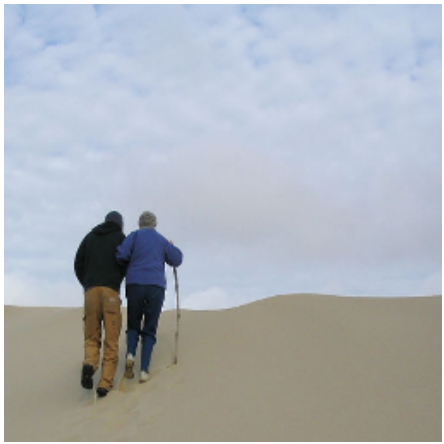 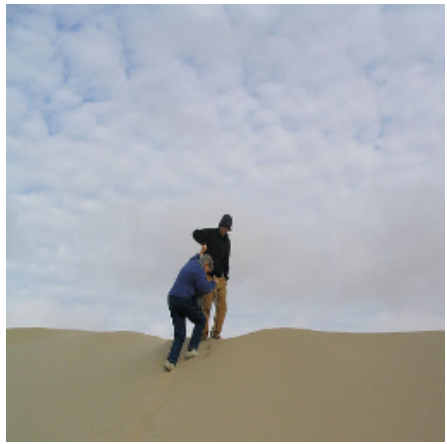

(a) A man and old woman with a walking stick climbing a sand dune arm in arm.

(b) One person helping another person up to the top of a mound of sand underneath a cloudy sky.

Fig. 1: Example images taken from the Flickr30K dataset along with their text descriptions. These two images and descriptions are semantically similar due to same scenes, but occur in different pairs; therefore, they are wrongly associated by networks relying on pairwise losses. However, in our network, the text encodings capture the semantic similarities between the two effectively.

pairwise loss functions usually ends up wrongly associating text neighbors of an image. The training objective of using these raw pairs is to adhere image and its text description in a pair and to ignore those descriptions which might be semantically similar but in different pairs, see Figure 1 where images (a) and (b) are semantically similar but occur in different pairs. Furthermore, multiple networks for each modality have a major drawback in terms of resource utilization.

In this paper we propose an image-text fusion that can capture semantic similarities between images and their text neighbors without suffering from typical image-text pairs problems i.e. mixing of semantically similar text descriptions because each text description has a different fused embedding. This leads to a single architecture which can be jointly trained to bridge gap between two modalities. We use centerloss [28] as a supervision signal to map images and associated text descriptions nearer to each other. We use InceptionResNet-V1 [29] as a baseline model. The network expects an image and encoded text descriptions and learns the similarity between them. We evaluate the network on two famous image retrieval datasets Flickr30k [30, 31] and MSCOCO [32]. Main contributions of our paper are the following:

- We show that image and text encoding can capture the semantic relationships between different modalities. To the best of our knowledge, this work is the first approach employing encoded text descriptions for cross modal retrieval.
- We show that a single network is equally capable of mapping two different modalities in a joint embedding space without having to use separate networks for each modality.

- Our single network is scaleable as compared to early fusion methods which employ single networks but are not scaleable.
- We extensively evaluate the network on two different datasets qualitatively and compute the Recall@K scores to show the effectiveness of our approach.

The rest of the paper is structured as follows: we explore the related literature in Section 2, proposed approach in Section 3, followed by dataset explanation in Section 4 and experiments in Section 5. Finally conclusions are in Section 6 followed by future work in Section 7.

## 2  Related Work

Several works in the field of multimodal representation learning have been proposed over recent years. Although each task is different from the other, the underlying principle is relatively same: to achieve semantic image-text multimodal representation. In this section we explore the related literature under different subsections.

### 2.1  Classical Approaches

One of the classical approaches towards image-text embedding is Canonical Correlation Analysis (CCA) [33]. The method finds linear projections that maximize the correlation between modalities. Works such as [34, 16] incorporate CCA to map representations of image and sentences to a common space. Although being a rather classical approach, the method is efficient enough. Recently, deep CCA has also been employed to the problem of obtaining a joint embedding for multimodal data [35]. However, the major drawback is that using CCA it is computationally expensive i.e. it requires to load all data into memory to compute the covariance score.

### 2.2  Deep Metric Learning Approaches

Deep metric learning based approaches have shown promising results on various computer vision tasks. The employing of metric learning to multimodal tasks requires within-view neighborhood preservation contraints which is explored in several works [36–38]. Triplet networks [39, 40] along with siamese networks [41–43] have been used to learn a similarity function between two modalities. However, most of these techniques [19] require separate networks for each modality which greatly increases the computational complexity of the whole process. Furthermore, these networks suffer from dramatic data expansion while creating sample pairs and triplets from the training set.

### 2.3  Ranking Supervision Signals

Many different multimodal approaches employ some kind of ranking loss function as a supervision signal. Works presented in [44, 2] employ a ranking loss which penalizes when incorrect description is ranked higher than the correct one. Similarly, the ranking loss can be employed in bi-directional fashion where the penalty is based on retrieval of both modalities.

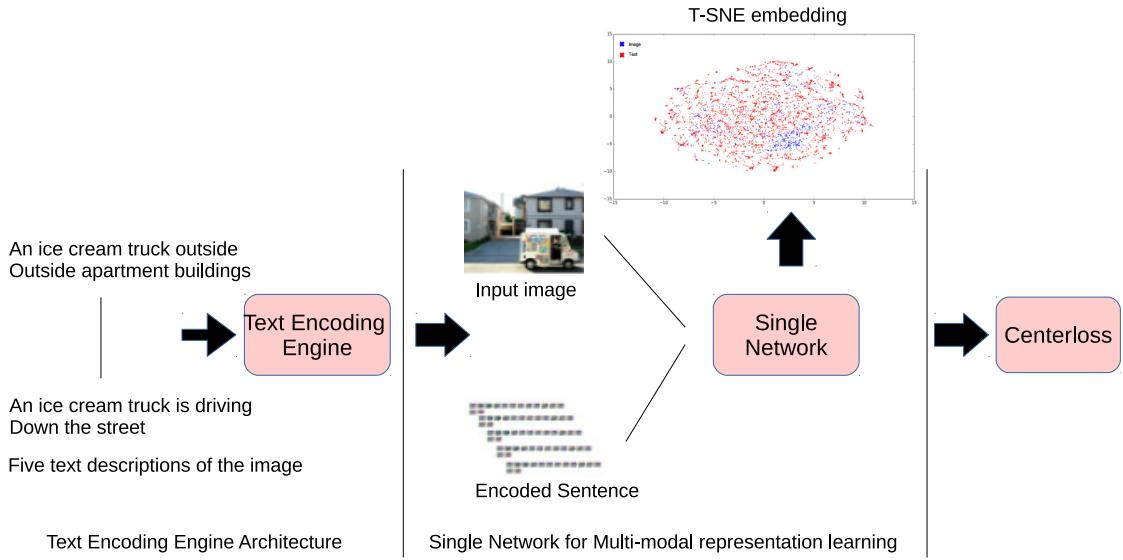

Fig. 2: Our single network (InceptionResNet-V1) based multimodal embedding architecture. The text encoding engine is a separate network which is used instead of standard Word2Vec. At top of the figure is t-SNE embedding of multimodal feature.

### 2.4   Classification Methods

Jointly representing multiple modalities on a common space can also be employed for classification purposes. Work in [45] employs classification loss along with two neural networks for both modalities (text and image) for zero-shot learning. Work in [46] employs attention-based mechanism to estimate the probability of a phrase over different region proposals in the image. In nearly every visual question answering (VQA) method, separate networks are trained for image and text; however, [47] treats the problem as a binary classification problem by using text as input and predicting whether or not an image-question-answer triplet is correct using softmax.

## 3   Proposed Approach

One of the core ideas of this paper is that the fusion of image and text can bridge the space between two modalities. We propose that with this fusion, the need for multiple networks for either modality becomes redundant since similar results can be achieved with a single network. The detailed proposed approach is presented in three different subsections: a) Image-Text Fusion, b) Single Network c) Loss Function. The Figure 2 visually explains the architecture of the network.

### 3.1 Image-Text Fusion

Semantics plays a crucial role to understand the meaning of a text document; humans can understand the semantics easily; however, for a computer semantics understanding is still a distant goal. Word2Vec word embedding [48] take one step towards mathematically representing the semantic relationships between word vectors since its objective function causes words that occur in similar contexts to have similar word embeddings. The work in [49] presented an encoding scheme by exploiting Word2Vec to reconstruct the semantics associated with a text document in an image, referred to as encoded image. We employ the encoding scheme presented in [49] to encode text descriptions and use them as input instead of standard raw text descriptions. The results presented verify that raw text features are not necessary to map descriptions to a similar embedding space with their respective images. With separate network for text descriptions not under consideration, a way is paved for computationally-inexpensive cross modal applications such as retrieval.

### 3.2 Single Network

We use the InceptionResNet-V1 as a baseline model, referred to as single network in this paper, see Figure 2. The network expects image along with associated encoded text descriptions as input. For example, in Flickr30K and MSCOCO datasets, there are five text description of an image so the input is an image along with five associated encoded text descriptions. Encoding text descriptions is similar to employing standard Word2Vec embedding for text descriptions before feeding them to the network. However, in this work, a single network effectively bridges image and associated text descriptions; due to this the need for multiple networks becomes expendable.

### 3.3 Loss Function

The network is trained using Adam Optimizer [50] with centerloss [28] as a supervision signal. Centerloss simultaneously learns a center for deep features of images in a mini-batch and penalizes the distances between the deep features and their corresponding neighbors, encoded text, and thus helps with neighborhood preserving constraints within each modality. Mathematically, centerloss is formulated as Eq. 1.

$$L_c = 1/2 \sum_{i=1}^{m} \| x_i - c_{y_i} \|_2^2 \qquad (1)$$

The $c_{y_i} \in \mathbb{R}^d$ corresponds to $y_i$th class center of deep features. As centerloss works to characterize intra-class variation, it effectively preserves the neighborhood structure. In this way those encoded text and images which are not semantically related do not occur in same neighborhood.

## 4 Dataset

For the retrieval task, we evaluate our approach on two very famous image-text retrieval datasets MSCOCO [32] and Flickr30K [30, 31]. Flickr30K contains 31,783 images that are collected from the Flickr website along with five captions for each image. We use 1000 images for testing and rest for training as described in [5]. MSCOCO contains 123,287 images, and each image is annotated with five captions. We use 1000 images for testing and the rest for training. Different papers employ different evaluation mechanisms on 5000 testing images of COCO dataset. Some present results on the entire 5k set, some choose 1k subsets while others average the results over the 5 folds. We present results on 1k set referred to as COCO-1k.

### 4.1 Data Augmentation

We perform multiple experiments with different data augmentation schemes. A standard setting consists of one image with five associated encoded text descriptions, referred to as $config-standard$. With further experiments on different augmentations of data, we find that increasing number of images at input do not help with the Recall@K scores. This is due to increasing semantic similarity between images due to frequent appearance of common objects in them e.g. person, vehicles etc. Since a person or a vehicle might appear in a large number of images, thus they tend to get wrongly associated. However, with distinct combinations of data, we find that increasing encoded text descriptions improve the Recall@K scores. In addition to image and its associated encoded text descriptions, we crop encoded text to a size of $227*227$ and feed them to the network. With this configuration, we have one image along with its horizontally flipped version and ten different encoded text descriptions, referred to as $config-2$. However, since the method does not require loading all the data in memory at once, bottlenecks are avoided. We report results on both configurations in Table 1 and 2

### 4.2 Training Details

The hyper parameters for the baseline version along with the supervision signal (centerloss) are fairly standard. For optimization, we employ Adam [50] because of its ability to adjust the learning rate during training. We use Adam's initial learning rate of 1e−3 and process 45 images in a batch. Furthermore, we resize the images to a size of $256*256$ before feeding into the network.

### 4.3 Evaluation Metric

For the evaluation of cross modal retrieval, we use same metric as described in [51] i.e. $R@K$ (read as Recall at K). $R@K$ is the percentage of queries in which the ground truth terms are one of the first K retrieved results. We employ the $R@1$, $R@5$ and $R@10$ which means that the percentage of queries in which the first 1, 5 and 10 items are ground truth terms respectively.

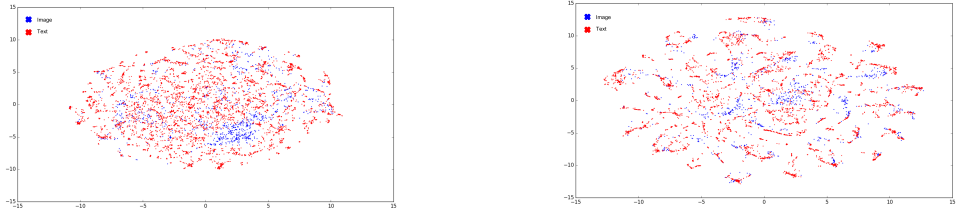

(a) t-SNE embedding for trained network on Flickr30k.

(b) t-SNE embedding for trained network on MSCOCO.

Fig. 3: t-SNE embedding with multimodal feature from image and text descriptions. (a) is the network trained with Flickr30K dataset and t-SNE embedding generated on its test set and (b) is the network trained with MSCOCO dataset and t-SNE embedding generated on its test set. (red: text, blue: image) (best viewed with color)

## 5 Experiments

This section presents experiments on two major datasets: Flickr30K and MSCOCO for cross modal retrieval. Images are resized to a standard size of $256 * 256$. The text embedding dimension is $128d$.

### 5.1 t-SNE

The Figure 3 is t-SNE [52] embedding result of multimodal features from Flickr30K's test dataset i.e. 1k images with five text descriptions for each image. Once the network is trained on the dataset, features of the test set are extracted from the model and are fed to t-SNE to visualize them, Figure 3(a), and thus image and its associated text are closer to each other. However, for the case of Figure 3(b), the network is trained on MSCOCO dataset and features are extracted from its test set i.e. 1k images with five text descriptions. Two different visualizations verify the claim that single network is capable of capturing the similarity and mapping image and encoded onto a shared space.The image and text encoding are overlapped in multimodal space and also distributed enough for being discriminated in retrieval. It means image and text encoding has similar distribution in multimodal space.

### 5.2 Retrieval Results

To evaluate the network for retrieval, we use Flickr30K and MSCOCO datasets. The datasets are split into train and test sets. We employ the Recall@K ($K = 1, 5, 10$) as an evaluation metric. To retrieve results, we take query image or text encoding and simply compute nearest neighbor (euclidean distance) between all images and sort results based on the distance. The first five distances correspond to Recall@K ($K = 5$) results and so on. The results on two different configurations ($config - standard$, $config - 2$) are reported in Table 1 and 2 along with comparison with the state-of-the-art methods. Table 3 is the visual illustration of Recall@K ($K = 5$); it shows various retrieval results from our single network based multimodal retrieval system.

## 5.3  Result Discussion

For benchmark of the network and comparison with other approaches, we did Recall@K evaluation for sentence-to-image and image-to-sentence retrieval. Compared to current state-of-the-art, our network's performance is comparatively low. We believe the major reason for these figures is due to the fact that Recall@K is based on whether query's pair appeared or not in retrieval result. So, even if retrieval result is semantically reasonable (Table 3) and if query's pair did not appear in the retrieval result, the Recall@K score can be considerably low. Furthermore, we recommend that retrieval systems should be evaluated on the efficiency of bridging the gap between multiple modalities and not on Recall@K which does not leverage on semantic relationships between modalities and rather focuses on Top-K retrieved results, which might not be correct every time. For example, the word *beach* is semantically related to every image which contains beach environment in it, however, if the word does not occur along with the image pair in Top K retrieved results, Recall@K comparatively gives low scores. Similar examples are presented in Figure 1. The work presented in [53] argued regarding Recall@K on similar grounds. Recall@K uses computable distance metric in input space to measure the distance between modalities. Secondly, we believe that since it does not compute a function that can accurately map input samples, the metric (Recall@K) suffers from inaccuracy in mapping semantically similar data points nearby on the manifold [54].

Table 1: Bidirectional retrieval Recall@K results on MSCOCO 1000-image test set (COCO-1k).

| Model | Image-to-Sentence | | | Sentence-to-Image | | |
|---|---|---|---|---|---|---|
| | R@1 | R@5 | R@10 | R@1 | R@5 | R@10 |
| DVSA [5] | 38.4 | 69.9 | 80.5 | 27.4 | 60.2 | 74.8 |
| HM-LSTM [55] | 43.9 | - | 87.8 | 36.1 | - | 86.7 |
| m-RNN-vgg [8] | 41.0 | 73.0 | 83.5 | 29.0 | 42.2 | 77.0 |
| Order-embedding [51] | 46.7 | - | 88.9 | 37.9 | - | 85.9 |
| m-CNN(ensemble) [56] | 42.8 | 73.1 | 84.1 | 32.6 | 68.6 | 82.8 |
| Structure Preserving [19] | 50.1 | 79.7 | 89.2 | 39.6 | 75.2 | 86.9 |
| TextCNN [53] | 13.6 | 39.6 | 54.6 | 10.3 | 35.5 | 55.5 |
| FV-HGLMM [53] | 14.3 | 40.5 | 55.8 | 12.7 | 39.0 | 57.2 |
| **Our Work** ($config - standard$) | − | − | − | − | − | − |
| **Our Work** ($config - 2$) | 12.50 | 30.10 | 40.0 | 11.18 | 32.22 | 45.76 |

Table 2: Bidirectional retrieval Recall@K results on Flickr30K 1000-image test set.

| Model | Image-to-Sentence | | | Sentence-to-Image | | |
|---|---|---|---|---|---|---|
| | R@1 | R@5 | R@10 | R@1 | R@5 | R@10 |
| mCNN(ensemble) [56] | 33.6 | 64.1 | 74.9 | 26.2 | 56.3 | 69.6 |
| m-RNN-vgg [8] | 35.4 | 63.8 | 73.7 | 22.8 | 50.7 | 63.1 |
| Deep CCA [57] | 27.9 | 56.9 | 68.2 | 26.8 | 52.9 | 66.9 |
| Structure Preserving [19] | 40.3 | 68.9 | 79.9 | 29.7 | 60.1 | 72.1 |
| Two-Way Nets [58] | 49.8 | 67.5 | - | 36.0 | 55.6 | - |
| **Our Work** ($config-standard$) | 10.5 | 26.2 | 36.8 | 8.2 | 22.82 | 32.0 |
| **Our Work** ($config-2$) | 14.5 | 33.0 | 42.6 | 10.5 | 26.74 | 37.2 |

## 6  Conclusions

We have proposed a novel approach for cross modal retrieval using image and text encoding. Our method employs single network to capture semantic relationships between multiple modalities. In contrast, until now all other methods exploit raw image text pairs and employ multiple networks for each modality. Furthermore, we emphasize how Recall@K is comparatively not a good metric to measure functionality and efficiency of multimodal retrieval systems. On basis of extensive experiments, we recommend that retrieval systems should be evaluated on the efficiency of bridging the gap between modalities and not on Recall@K which does not leverage on semantic relationships between modalities.

## 7  Future Work

Our work can be extended to other multimodal representation learning (e.g. sound-image, video-text, etc). So, our method's future work will be extending this method to other multimodal cases. We would also want to work on finding a better metric based on dimensionality reduction phenomenon [54] to evaluate cross modal retrieval systems. Furthermore, we would like to explore different data augmentation to find the most effective configuration which helps in semantically aligning images and its defining descriptions in a neighborhood. Research towards different encoding schemes to exploit the semantic dependencies between sentences is another possibility.

Table 3: Image to sentence retrieval result (top 5) from the Flickr30K dataset. Query and retrieved sentences are semantically related but not necessarily in pairs.

| Image | Retrieved Sentences |
| --- | --- |
| 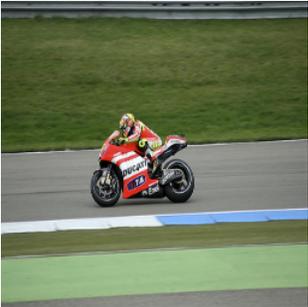 | – A motorcycle racer in red and yellow leathers is riding a red and white Ducati sport bike and leaned over in a right hand turn on a track.<br>– A man wearing red and yellow riding a red sport bike driving on a racetrack.<br>– A person rides a motorcycle on a racetrack.<br>– A man in a red and white helmet and race gear is riding a recumbent bicycle.<br>– A man on a speed motorcycle rides on a track. |
| 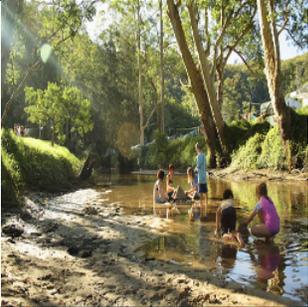 | – Some children are playing in a woodland stream.<br>– A group of children playing in a stream.<br>– Five people enjoying time in nature.<br>– Several people are congregating near an area with a tree, lots of bicycles and some debris, including old tires.<br>– A group of children play in the muddy water near the woods. |
| 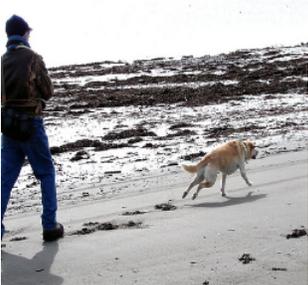 | – A man walking behind a running dog on the beach.<br>– A woman leaps across the beach.<br>– Two men are holding a net while standing on the beach.<br>– Two dogs playing by the shore.<br>– A man and a dog on the beach. |
| 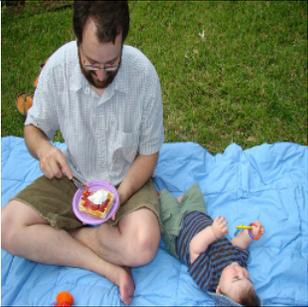 | – A man sitting on a blanket with his kid while eating a waffle with whip cream and fruit.<br>– A man in sunglasses sits in a chair holding a metal pan, surrounded by a lot of stuff.<br>– A man is sitting on a chair holding a large stuffed animal.<br>– A young man using a notepad and pen to diagram an image. |
| 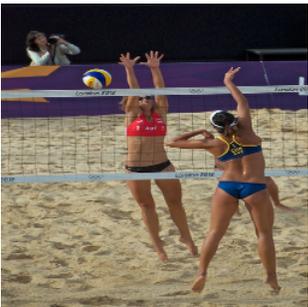 | – Two women in bathing suits play beach volleyball , one is serving and the other is preparing to hit the ball.<br>– The players are on the height of their energies both aiming to win on a beach volleyball game.<br>– Two women are on the tennis court wearing skirts.<br>– Two girls playing volleyball on the beach.<br>– 3 ladies with numbers on their shirts running. |

14        Authors Suppressed Due to Excessive Length